\documentclass[runningheads]{llncs}
\usepackage{graphicx}
\usepackage{subcaption}
\begin{document}
\title{ A Pretrained DenseNet Encoder for Brain Tumor Segmentation. }

\author{Jean Stawiaski \inst{1} }
\authorrunning{J. Stawiaski}
%
\institute{Stryker Corporation, Navigation. Freiburg im Breisgau, Germany.
\email{jean.stawiaski@stryker.com}}
\maketitle              
\begin{abstract}
This article presents a convolutional neural network for the automatic segmentation of brain tumors in multimodal 3D MR images based on a U-net architecture. We evaluate the use of a densely connected convolutional network encoder (DenseNet) which was pretrained on the ImageNet data set. We detail two network architectures that can take into account multiple 3D images as inputs. This work aims to identify if a generic pretrained network can be used for very specific medical applications where the target data differ both in the number of spatial dimensions as well as in the number of inputs channels. Moreover in order to regularize this transfer learning task we only train the decoder part of the U-net architecture. We evaluate the effectiveness of the proposed approach on the BRATS 2018 segmentation challenge \cite{brats0,brats1,brats2,brats3,brats4} where we obtained dice scores of 0.79, 0.90, 0.85 and 95\% Hausdorff distance of 2.9mm, 3.95mm, and 6.48mm  for enhanced tumor core, whole tumor and tumor core respectively on the validation set. This scores degrades to 0.77, 0.88, 0.78 and 95\% Hausdorff distance of 3.6mm, 5.72mm, and 5.83mm on the testing set \cite{brats0}.

\end{abstract}

\section{Introduction}

\begin{figure}[httb]
    \centering
    \includegraphics[width=\textwidth]{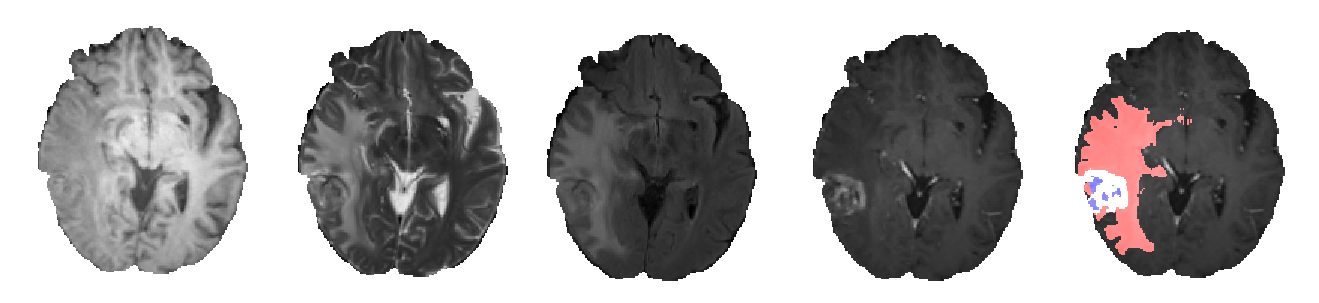}
    \caption{Example of images from the BRATS 2018 dataset. From left to right: T1 image, T2 image: the whole tumor and its core are visible, T2 FLAIR image: discarding the cerebrospinal fluid signal from the T2 image highlights the tumor region only, T1ce: contrast injection permits to visualize the enhancing part of the tumor as well as the necrotic part. Finally the expected segmentation result is overlaid on the T1ce image. The edema is shown in red, the enhancing part in white and the necrotic part of the tumor is shown in blue. }
    \label{bratsex}
\end{figure}

Automatic segmentation of brain tumor structures has a great potential for surgical planning and intraoperative surgical resection guidance. Automatic segmentation still poses many challenges because of the variability of appearances and sizes of the tumors. Moreover the differences in the image acquisition protocols, the inhomogeneity of the magnetic field and partial volume effects have also a great impact on the image quality obtained from routinely acquired 3D MR images. However brain gliomas can be well detected using modern magnetic resonance imaging. The whole tumor is particularly visible in T2-FLAIR, the tumor core is visible in T2 and the enhancing tumor structures as well as the necrotic parts can be visualized using contrast enhanced T1 scans. An example is illustrated in figure \ref{bratsex}. \\

In the recent years, deep neural networks have shown to provide state-of-the-art performance for various challenging image segmentation and classification problems \cite{FCN,FCN-CRF,segnet,dilnet,Deconv}. Medical image segmentation problems have also been successfully tackled by such approaches \cite{UNET,VNET,deepsuper2,deepsuper3,cascade}. However training deep neural networks can still be challenging in the case of a limited number of training data. In such situations it is often necessary to limit the complexity and the expressivity of the network. It has been observed that initializing weights of a convolutional network that has been pretrained on a large data set improves its accuracy on specific tasks where a limited number of training data is available \cite{ternaus}. We evaluate in this work the accuracy of a U-net architecture \cite{UNET,VNET} where the encoder is a densely connected convolutional network \cite{densenet} which has been pretrained on the ImageNet data set \cite{ImageNet}. We study an extreme case of transfer learning where we fix the weights of the pretrained DenseNet encoder. Moreover we consider a segmentation problem where the input data dimensionality does not match the native input dimensions of the pretrained network. We will thus make use of a fixed pretrained network trained on 2D color images in order to segment 3D multimodal medical images. We will see that fixing the weights of the encoder is a simple but effective way to regularize the segmentation results.

\section{Method}

This section details the proposed network architectures, the loss function used to train the network as well as the training data preparation.

\subsection{Convolutional Neural Network Architectures}

\begin{figure}[httb]
    \centering
    \includegraphics[width=0.75\textwidth]{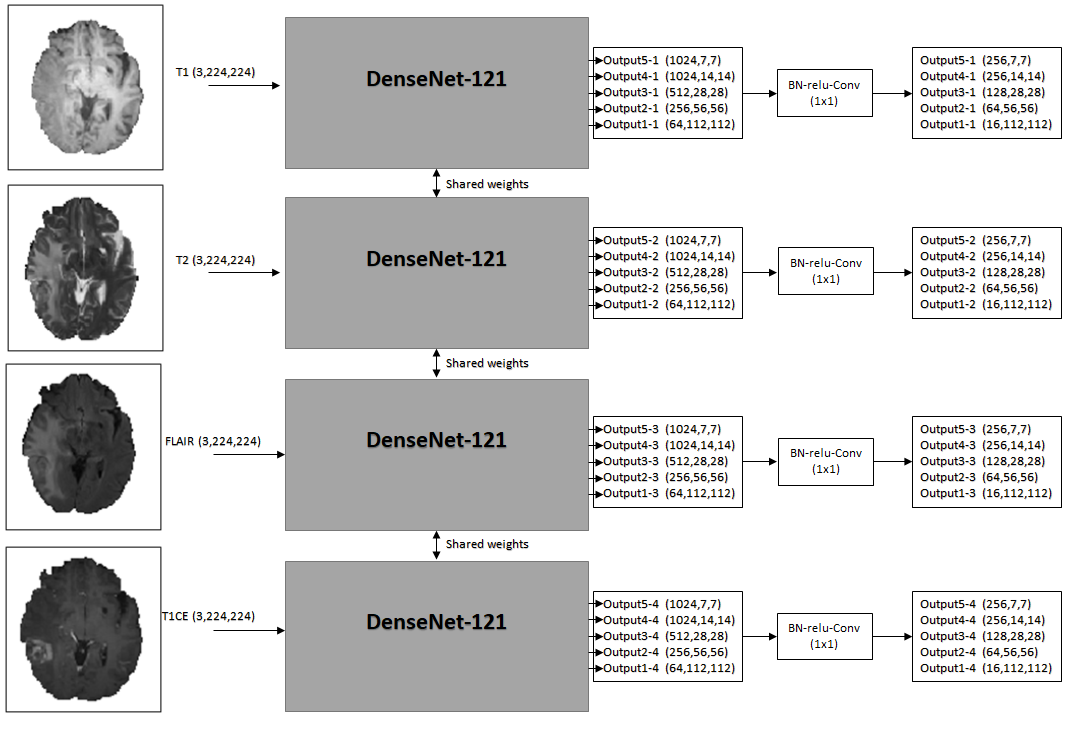}
    \caption{Encoder architecture (M2). The network processes the different input image modality with the same encoder, a DenseNet composed of 121 layers. Intermediate layers of the encoder are used to feed the decoder network.  }
    \label{encoder1}
\end{figure}

The network processes 2D images of size (224,224) pixels containing three channels. An input image is composed of three successive slices of the input volume along one of the three anatomical orientations: either along the coronal, the sagittal or the transverse plane. We use a pretrained network that has been designed to take a single 2D color image as input. In order to be able to process multi modal inputs, we have designed two distinct architectures:

\begin{itemize}

\item the first solution (M1) consists in removing the stem of the original DenseNet and only make use of the following convolutional layers which input is a tensor of size (64,112,112). This architecture is illustrated in figure \ref{archi}. The proposed network is composed of a "precoder" which produces an adequate high dimensional input tensor for the pretrained network. This architecture is illustrated in figure \ref{precoder}. It processes independently each input images and concatenates the resulting tensors. This approach is very flexible and could take as input an image of any dimensions.\\

\begin{figure}[httb]
    \centering
    \includegraphics[width=0.75\textwidth]{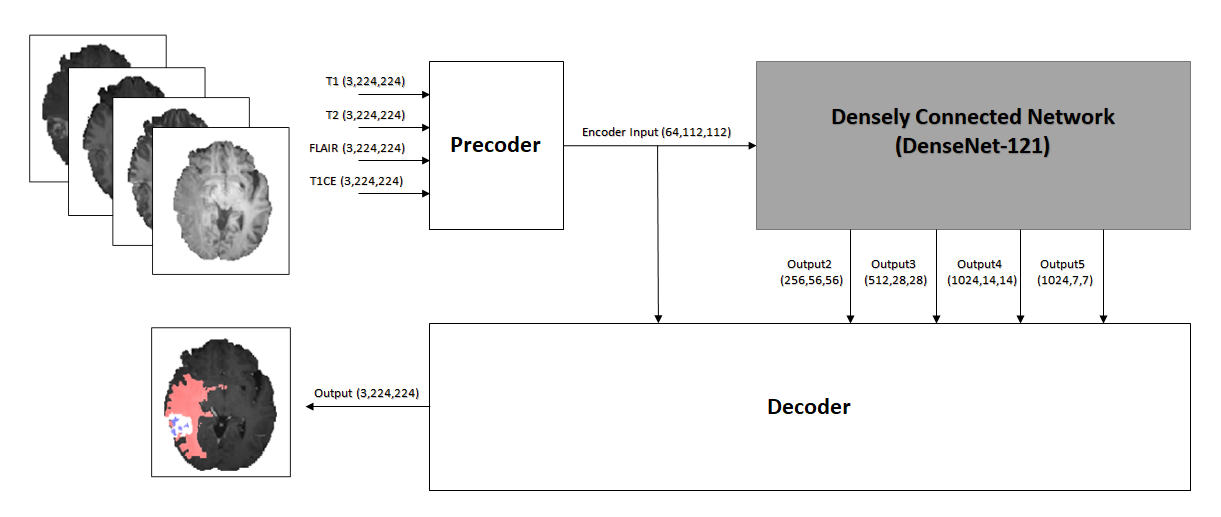}
    \caption{Network architecture (M1). The network is composed of a "precoder" producing a high order tensor which is fed to a pretrained densely connected convolutional network. Several intermediate layers are then used to reconstruct a high resolution segmentation map. }
    \label{archi}
\end{figure}

\begin{figure}[httb]
    \centering
    \includegraphics[scale=0.3]{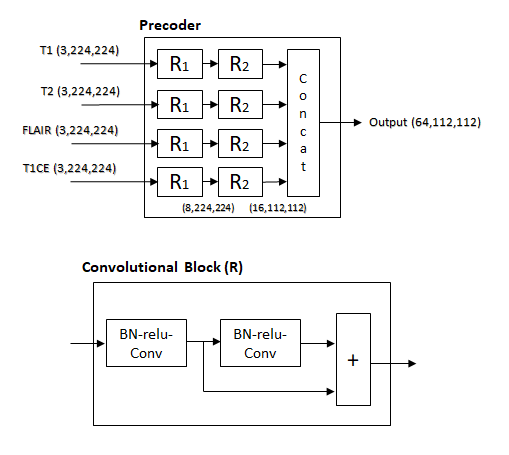}
    \caption{Precoder architecture (M1). The precoder architecture process independently the input images by a sequence of multiple residual blocks (R1, R2) and concatenates the resulting output tensors. A residual block (R) is also illustrated. All convolution operations are computed with (3x3) kernels. }
    \label{precoder}
\end{figure}

\item the second solution (M2) consists in evaluating the different input modality separately through the original DenseNet encoder. Each input image modality is processed with the same encoder which shares its weights across the different modalities. Outputs at different scales are then concatenated and fed to the decoder. This architecture is illustrated in figure \ref{encoder1} and \ref{decoder}. This architecture does not permit to vary the number of input slices but has the advantage to fully leverage the original DenseNet weights.

\end{itemize}

For both architectures, the decoder consists in upsampling a low resolution layer, concatenate it with a higher resolution layer before applying a sequence of convolution operations. The first convolutional layer reduces the number of input channels by applying a (1x1) convolution. Following layers are composed of spatial (3x3) convolutions with residual connections.\\

\begin{figure}[httb]
    \centering
    \includegraphics[width=\textwidth]{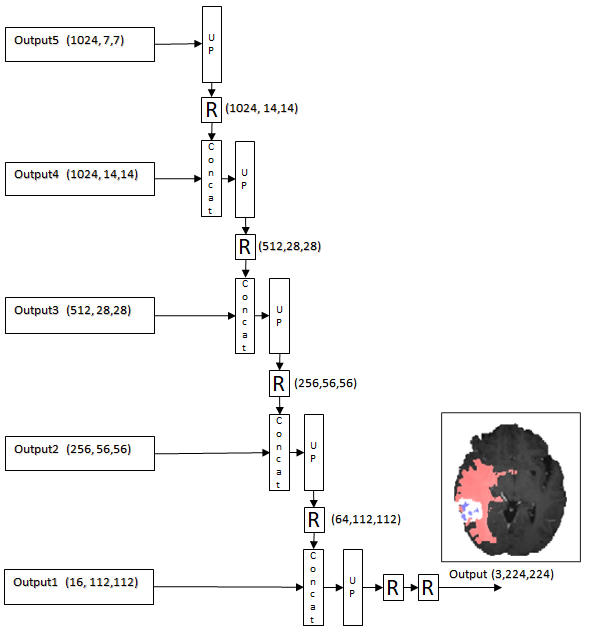}
    \caption{Decoder architecture of the first model (M1). The decoder consists in a sequence of upsampling and residual convolution operations in order to produce a high resolution segmentation map. }
    \label{decoder}
\end{figure}

\begin{figure}[httb]
    \centering
    \includegraphics[width=\textwidth]{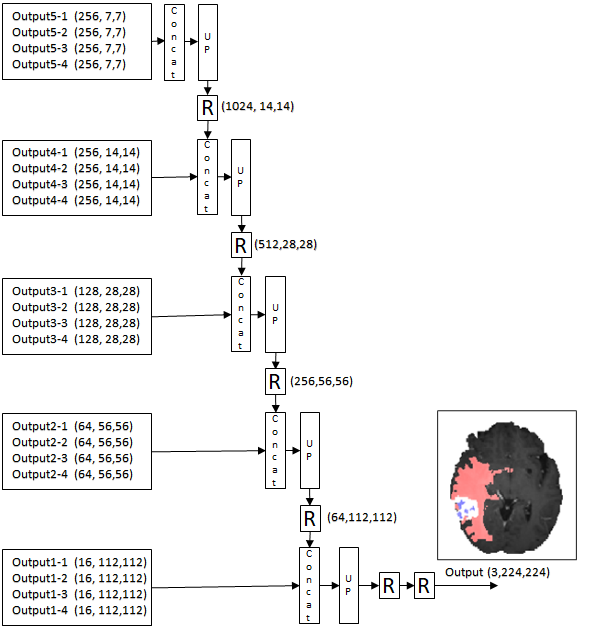}
    \caption{Decoder architecture of the second model (M2). The decoder concatenates the encoding layers of each modalities. The segmentation is produced with a sequence of upsampling and convolution operations. }
    \label{decoder}
\end{figure}

 \vspace{5mm}

We give here additional details about the network architectures:
\begin{itemize}

\item each sample 3D image $y$ is normalised so that voxels values falls in the interval $[0,1]$.

\item batch normalisation is performed after each convolutional layer using a running mean and standard deviation computed on 5000 samples:

\item each layer is composed of residual connections as illustrated in figure \ref{decoder},

\item the activation function used in the network is a rectified linear unit,

\item convolutions are computed using reflective border padding,

\item upsampling is performed by nearest neighbor interpolation.

\end{itemize}

\clearpage

\subsection{Training}

We used the BRATS 2018 training and validation sets for our experiments \cite{brats1,brats2,brats3,brats4}. The training set contains 285 patients (210 high grade gliomas and 75 low grade gliomas). The BRATS 2018 validation set contains 66 patients with brain tumors of unknown grade with unknown ground truth segmentations. Each patient contains four modalities: T1, T1 with contrast enhancement, T2 and T2 FLAIR. The aim of this experiment is to segment automatically the whole tumor, the tumor core and the tumor enhancing parts. Note that the outputs of our neural network corresponds directly to the probability that a pixel belongs to a tumor, the core of a tumor and the enhancing part of the tumor. The last layer of the proposed architecture is thus composed of three independent (1x1) convolutional layers because we directly model the problem as a multi-label segmentation problem where a pixel can be assigned to multiple classes. Note that only weights of the "precoder" and the decoder are learned. Original weights of the pretrained DenseNet-121 stay fixed during the training procedure.\\

The network produces a segmentation maps by minimizing a loss function defined as the combination of the mean cross entropy (mce) and the mean Dice coefficients (dce) between the ground truth class probabilities and the network estimates:
\begin{equation}
ce = \sum_k \Big( \frac{-1}{n} \sum_i y_i^k log(p_i^k) \Big) \;
\end{equation}
where $y_i^k$ and $p_i^k$ represent respectively the ground truth probability and the network estimate for the class $k$ at location $i$.

\begin{equation}
dce = \sum_{k \neq 0}  \Big( 1.0 - \frac{1}{n} \Big( \frac{ 2 \sum_i p_i^k y_i^k + \epsilon }{ \sum_i (p_i^k)  + \sum_i (y_i^k) + \epsilon  } \Big) \Big) \; .
\end{equation}

\begin{figure}[httb]
    \centering
    \includegraphics[width=\textwidth]{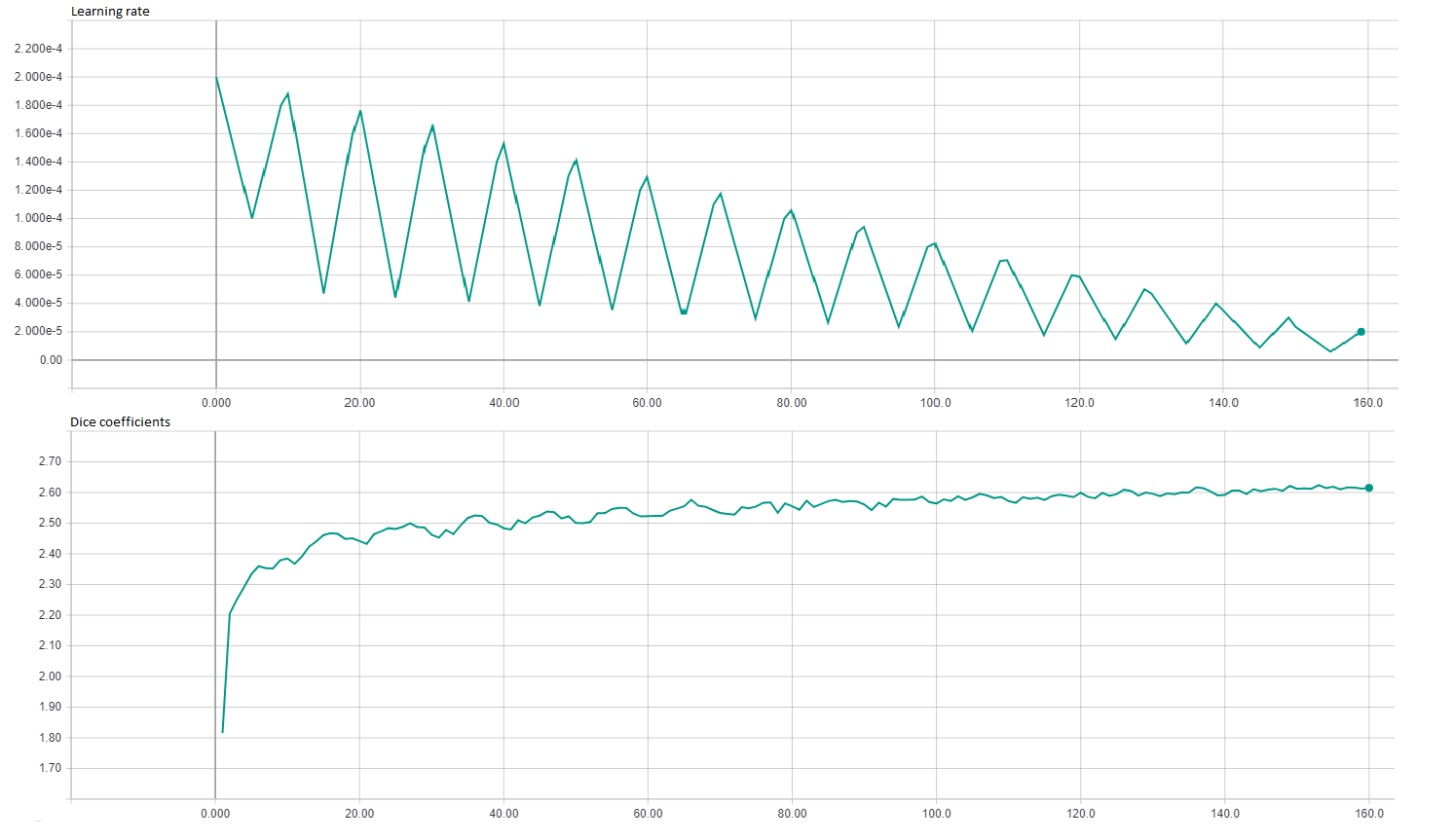}
    \caption{Network training. Illustration of the cyclic learning rate schedule (top). Evolution of the sum of the dice coefficients of the three classes during training (bottom). }
    \label{training}
\end{figure}

Note that we exclude the background class for the computation of the dice coefficient. The network is implemented using Microsoft CNTK \footnote{\url{https://www.microsoft.com/en-us/cognitive-toolkit/}}. We use stochastic gradient descent with momentum to train the network and L2 weights regularization. We use a cyclic learning rate where the learning rate varies from 0.0002 to 0.00005. An example of the evolution of the accuracy and the learning rate is illustrated in figure \ref{training}. We train the network for 160 epochs on a Nvidia GeForce GTX 1080 GPU. A full epoch consists in analyzing all images of the BRATS training data set and extracting 20 2D random samples from the 3D MR volumes.

\subsection{Testing}

Segmentation results are obtained by evaluating the network along slices extracted from the three anatomical orientations and averaging the results. A segmentation map is then obtained by assigning to each voxel the label having the maximum probability among the three classes: tumor, tumor core or enhancing tumor. Finally connected components composed of less than 100 voxels are removed. We are not making use of test time image augmentation or ensembling methods.\\

\section{Results}

We uploaded our segmentation results to the BRATS 2018 server \footnote{\url{https://www.cbica.upenn.edu/BraTS18/lboardValidation.html} } which evaluates the segmentation and provides quantitative measurements in terms of Dice scores, sensitivity, specificity and Hausdorff distances of enhanced tumor core, whole tumor, and tumor core. Results of the BRATS 2018 validation phase are presented in Table 1. The validation phase is composed of 66 datasets with unknown ground truth segmentations. \\

\begin{center}
  \begin{tabular}{ | l | l | l | l | l | l | l | }
    \hline
      & Dice ET & Dice WT & Dice TC & Dist. ET & Dist. WT & Dist. TC \\ \hline
    \textbf{Mean M1}  &  0.768         &          0.892  &  0.815         &  3.85         &  4.85         &  7.56       \\
    \textbf{Mean M2}  & \textbf{0.792} & \textbf{ 0.899} & \textbf{0.847} & \textbf{2.90} & \textbf{3.95} & \textbf{6.48} \\
    \hline
    \textbf{StdDev M1 }  &  0.241         &          0.065  &  0.187         &  5.43         &  4.28         &  12.56  \\
    \textbf{StdDev M2}   & \textbf{0.223} &  \textbf{0.074} & \textbf{0.130} & \textbf{3.59} & \textbf{3.38} & \textbf{12.06} \\
    \hline
    \textbf{Median M1 }  &         0.849   &  0.905         &         0.889  &         2.23  &        3.67   &  3.74  \\
    \textbf{Median M2 }  &  \textbf{0.864} & \textbf{0.919} & \textbf{0.891} & \textbf{1.73} & \textbf{3.08} & \textbf{3.30} \\
    \hline
    \textbf{25\% quantile M1 }  & \textbf{0.792} &   0.881        &  0.758         &  1.68         & \textbf{2.23} & \textbf{2} \\
    \textbf{25\% quantile M2 }  &  0.789         &  \textbf{0.890}& \textbf{0.796} & \textbf{1.41} & \textbf{2.23} & \textbf{2} \\
    \hline
    \textbf{75\% quantile M1 }  &  0.888          &         0.933 &  0.930         &  3.16         &  5.65         &         8.71  \\
    \textbf{75\% quantile M2}   & \textbf{0.906}  & \textbf{0.939}& \textbf{0.932} & \textbf{2.82} & \textbf{4.41} & \textbf{6.65} \\
    \hline
  \end{tabular}
  \captionof{table}{BRATS 2018 Validation scores, dice coefficients and the 95\% Hausdorff distances in mm. Our results corresponds to the team name "Stryker". (M1) results corresponds to the precoder approach, (M2) corresponds to the direct use of a fixed pretrained DenseNet-121. }
\end{center}

Results of the BRATS 2018 testing phase are presented in Table 2. The testing phase is composed of 191 datasets with unknown ground truth segmentations.\\

\begin{center}
  \begin{tabular}{ | l | l | l | l | l | l | l | }
    \hline
      & Dice ET & Dice WT & Dice TC & Dist. ET & Dist. WT & Dist. TC \\ \hline
    \textbf{Mean M2}          & \textbf{0.776} &  \textbf{0.878} & \textbf{0.786} & \textbf{3.63} & \textbf{5.72} & \textbf{5.83} \\
    \hline
    \textbf{StdDev M2}        & 0.223 &  0.104 & 0.257 & 5.29 & 7.31 & 7.93 \\
    \hline
    \textbf{Median M2 }       & 0.828 &  0.908 & 0.891 & 2.23  & 3.60 & 3.46 \\
    \hline
    \textbf{25\% quantile M2 }  & 0.749 &  0.857 & 0.796 & 1.41  & 2.23 & 2.1  \\
    \hline
    \textbf {75\% quantile M2}  & 0.895 &  0.935 & 0.924 & 3.0   & 6.08 & 6.13 \\
    \hline
  \end{tabular}
  \captionof{table}{BRATS 2018 Testing scores, dice coefficients and the 95\% Hausdorff distances in mm.  }
\end{center}

\begin{figure}[httb]
    \centering
    \includegraphics[width=\textwidth]{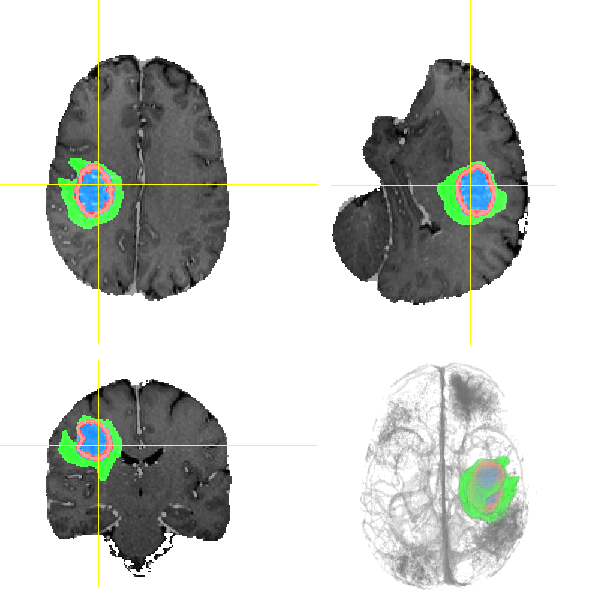}
    \caption{Segmentation result obtained on a image of the testing data. }
    \label{train}
\end{figure}

\section{Discussion}

The validation and testing results obtained on the BRATS segmentation challenge show that the proposed approaches are indeed efficient. Despite the fact that the used encoder has been trained on natural color images, it turns out that the learned features can be leveraged for a large class of applications including segmentation of medical images. Using a fixed encoder is thus an effective way to regularize the neural network. Note that we did not make use of advanced image augmentations or ensembling methods. The two approaches produce comparable results and have both advantages and drawbacks. The model (M1) is more versatile since it can use any number of input modalities (channels) and any number of spatial dimensions. However current experiments shows that the model (M2), despite its simplicity, produces slightly better results. A major limitation of the proposed approach is the lack of 3D spatial consistency.

\section{Conclusion}

We have studied an extreme version of transfer learning by using a fixed pretrained network trained on 2D color images for segmenting 3D multi modal medical images.
We have presented two simple approaches for leveraging pretrained networks in order to perform automatic brain tumor segmentation. We obtained competitive scores on the BRATS 2018 segmentation challenge \footnote{\url{https://www.cbica.upenn.edu/BraTS17/lboardValidation.html}}. Future work will concentrate on several possible improvements by additionally fine tuning the pretrained encoder. A fixed large expressive 2D neural network is thus an interesting alternative to a relative small task specific 3D neural networks.\\

%
%
%
%

\end{document}